\documentclass{article}
\usepackage{spconf,amsmath,amssymb,graphicx}
\usepackage{subfig}
\usepackage{bm}
\usepackage{color}
\usepackage{algorithm}
\usepackage{algorithmic}
\usepackage{caption}
\usepackage{pgfplots}
\usepackage{hyperref}
\usepackage{caption}

\newcommand{\wrt}{\text{w.\,r.\,t.}~}

\def\mat#1{\ensuremath{\bm{\mathit{#1}}}}
\def\vec#1{{\ensuremath{\bm{\mathit{#1}}}}}

\newcommand{\NoiseStd}{\sigma_{\text{noise}}}
\newcommand{\PriorStd}{\sigma_{\text{prior}}}

\newcommand{\FrameIdx}[2]{#1^{(#2)}}					
\newcommand{\IterationIdx}[2]{#1^{#2}}			

\newcommand{\HRSym}{x}
\newcommand{\HR}{\vec{\HRSym}}													
\newcommand{\LRSym}{y}
\newcommand{\LR}{\vec{\LRSym}}													
\newcommand{\LRFrame}[1]{\FrameIdx{\LR}{#1}}						

\newcommand{\SystemMat}{\mat{W}}

\newcommand{\MotionMat}{\mat{M}}

\newcommand{\BlurMat}{\mat{H}}
\newcommand{\SamplingMat}{\mat{D}}

\newcommand{\PSFWidth}{\sigma_{\text{PSF}}}

\newcommand{\MotionParams}{\bm{\theta}}

\newcommand{\RegTerm}[2][]{R_{\text{#1}}(#2)}

\newcommand{\RegWeight}[1][]{\lambda_{\text{#1}}}

\newcommand{\WeightsAFun}{\alpha}
\newcommand{\WeightsAMat}{\mat{A}}

\newcommand{\WeightsBFun}{\beta}
\newcommand{\WeightsBMat}{\mat{B}}

\pgfplotsset{compat=newest}
\title{Confidence-Aware Levenberg-Marquardt Optimization for Joint Motion Estimation and Super-Resolution}
\name{Cosmin Bercea, Andreas Maier, and Thomas K\"ohler}
\address{Pattern Recognition Lab, Friedrich-Alexander-Universit\"at Erlangen-N\"urnberg, Erlangen, Germany
}
\begin{document}
%
\maketitle
\begin{abstract}
Motion estimation across low-resolution frames and the reconstruction of high-resolution images are two coupled subproblems of multi-frame super-resolution. This paper introduces a new joint optimization approach for motion estimation and image reconstruction to address this interdependence. Our method is formulated via non-linear least squares optimization and combines two principles of robust super-resolution. First, to enhance the robustness of the joint estimation, we propose a confidence-aware energy minimization framework augmented with sparse regularization. Second, we develop a tailor-made Levenberg-Marquardt iteration scheme to jointly estimate motion parameters and the high-resolution image along with the corresponding model confidence parameters. Our experiments on simulated and real images confirm that the proposed approach outperforms decoupled motion estimation and image reconstruction as well as related state-of-the-art joint estimation algorithms.             
\end{abstract}

\section{Introduction}
\label{sec:Introduction}

In digital imaging, super-resolution (SR) refers to a class of computational techniques to improve the spatial resolution of an imaging system. Over the past years, a variety of algorithms have been introduced that approach SR either from a single-frame \cite{Freeman2000a} or a multi-frame perspective. Multi-frame methods exploit subpixel displacements across a sequence of low-resolution (LR) frames to reconstruct a high-resolution (HR) image \cite{Park2003}. This process can be considered as the combination of an initial motion estimation followed by image reconstruction. If motion is estimated properly, this approach holds the potential to overcome aliasing artifacts due to undersampling. Unfortunately, in the presence of inaccurate motion estimation, its actual performance deteriorates. 

Principally, SR can be approached by two basic methods:\\ 
1) The \textsl{two-stage} approach \cite{Elad1997} considers motion estimation and image reconstruction as independent subproblems and employs image registration on LR frames to determine subpixel motion that is directly utilized for image reconstruction. The main drawback of this approach is the limited accuracy of motion estimation on LR frames \cite{Vandewalle2006}, which leads to a poor robustness of the overall process. State-of-the-art methods aim at compensating inaccurate motion estimation explicitly by tailor-made outlier detection techniques \cite{Zhao2002} or implicitly by robust optimization \cite{Zomet2001}. In this context, SR has been studied with outlier-insensitive observation models including robust error norms \cite{Farsiu2004a,Vrigkas2014} or, more recently, by space variant Bayesian models \cite{Kohler2016}. Another trend are hybrid methods that combine single-frame and multi-frame SR to locally enhance the image reconstruction \cite{Batz2015}.\\ 
2) \textsl{Joint} estimation as a complementary class of algorithms considers motion estimation and image reconstruction as two coupled subproblems. This aims at simultaneously estimating motion parameters and the associated HR image to enhance the accuracy of both tasks. For this purpose, alternating minimization regarding both subproblems \cite{Hardie1997,Liu2011} is a common optimization scheme. More recently, different Bayesian formulations including marginalization over the HR image \cite{Tipping2003} or the motion parameters \cite{Pickup2007}, as well as variational inference \cite{Babacan2011} have been developed to overcome the poor convergence of alternating minimization. Other state-of-the-art approaches related to our work include joint Gauss-Newton iterations \cite{He2007,Zhang2012} or linear programming \cite{Yap2009}.
\begin{figure}[!t]
	\centering
	\subfloat[Original]{\includegraphics[width=0.152\textwidth]{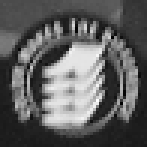}}~
	\subfloat[Method in \cite{He2007}]{\includegraphics[width=0.152\textwidth]{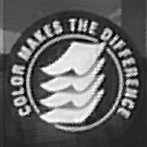}}~
	\subfloat[Our method]{\includegraphics[width=0.152\textwidth]{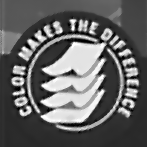}}
	\caption{Super-resolution under inaccurate motion estimation including model outliers related to a camera zoom.}
	\label{fig:introExample}
\end{figure}
Compared to the two-stage approach, these schemes are able to correct inaccurate motion estimation up to a certain degree. However, one of their main limitations is that they are often based on simplified models and are sensitive to outliers in the image formation. Fig.\,\ref{fig:introExample} depicts this issue on an example dataset affected by outliers in the motion model with a comparison of the Gauss-Newton algorithm in \cite{He2007} to our proposed method.

This paper proposes a new algorithm for joint motion estimation and image reconstruction. As the main contribution, we propose non-linear least squares optimization regarding the motion parameters and the HR image based on a confidence-aware formulation of both subproblems. In addition, we develop a tailor-made Levenberg-Marquardt iteration scheme to jointly estimate motion parameters, the HR image and model confidence parameters. MATLAB code of this method is available on our webpage as part of our super-resolution toolbox\footnote{\scriptsize \hyperref[www5.cs.fau.de/research/software/multi-frame-super-resolution-toolbox]{www5.cs.fau.de/research/software/multi-frame-super-resolution-toolbox}}. 

\section{Confidence-Aware Joint Motion Estimation and Super-Resolution}
\label{sec:ConfidenceAwareJointMotionEstimationAndSuperResolution}

\indent
This section introduces our joint motion estimation and SR algorithm. First, Section \ref{sec:ConfidenceAwareModeling} presents the confidence-aware formulation for our method. Then, in Section \ref{sec:NonLinearLeastSquaresEstimation} and Section \ref{sec:LevenbergMarquardtIterations}, we develop non-linear least-squares estimation for this model and the proposed Levenberg-Marquardt optimization.

\subsection{Confidence-Aware Modeling}
\label{sec:ConfidenceAwareModeling}

We describe the formation of a set of LR frames assembled to $\LR = (\LRFrame{1}\,^\top, \ldots, \LRFrame{K}\,^\top)^\top$, $\LRFrame{k} \in \mathbb{R}^M$ from the HR image $\HR \in \mathbb{R}^N$ by: 
\begin{equation}
	\label{eqn:imageFormationModel}
	\LR = \SystemMat(\MotionParams)\HR + \epsilon,
\end{equation}
where $\SystemMat(\MotionParams) = \SamplingMat\BlurMat\MotionMat(\MotionParams)$ denotes the system matrix and $\epsilon$ is additive noise \cite{Elad1997}. We decompose $\SystemMat(\MotionParams)$ into $\SamplingMat$ to model subsampling, $\BlurMat$ to model space invariant blur related to the camera point spread function (PSF) and $\MotionMat(\MotionParams)$ to describe subpixel motion of the different frames according to the motion parameters $\MotionParams = (\FrameIdx{\MotionParams}{1},\ldots, \FrameIdx{\MotionParams}{K})^\top$ relative to $\HR$. In this paper, we limit ourselves to a rigid motion model described by the parameters $\FrameIdx{\MotionParams}{k} = (\cos(\varphi), \sin(\varphi), t_u\textrm{, } t_v)^\top$ with rotation angle $\varphi$ and translation $\vec{t} = (t_u\textrm{, } t_v)^\top$ for the frame $\LRFrame{k}$.

We propose to couple SR and motion estimation in a joint optimization framework. Accordingly, we infer the unknown HR image $\HR$ and the motion parameters $\MotionParams$ as the minimum of: 
\begin{equation}
	\label{eqn:jointEnergyFunction}
	E(\HR, \MotionParams) 
	= \left(\LR - \SystemMat(\MotionParams) \HR\right)^\top 
	\WeightsBMat \left(\LR - \SystemMat(\MotionParams) \HR\right) + \RegWeight \cdot \RegTerm{\HR},
\end{equation}
where $\RegTerm{\HR}$ is a regularizer with weight $\RegWeight \geq 0$. To make the joint estimation in \eqref{eqn:jointEnergyFunction} robust regarding outliers, we adopt the space variant observation model developed in our prior work \cite{Kohler2016} that is defined by the confidence map $\WeightsBMat = \text{diag}(\WeightsBFun _1, \ldots, \WeightsBFun_{KM})$ to weight the influence of the individual observations by adaptive weights $\WeightsBFun_i \in [0; 1]$. Moreover, we use weighted bilateral total variation (WBTV) \cite{Kohler2016} as a sparse prior for edge preserving regularization. We define $\RegTerm{\HR}$ by the sparsifying transform $\vec{S}$ according to:
\begin{equation}
	\label{eqn:RegTerm}
	\RegTerm{\HR} = \left|\left| \WeightsAMat \vec{S} \HR \right|\right|_1 
	= \sum_{l = -P}^P \sum_{m = -P}^P \left|\left| \WeightsAMat^{l,m} \vec{S}^{l,m} \HR \right|\right|_1,
\end{equation}
where $\vec{S}^{l,m} = \alpha_0^{|l| + |m|} (\vec{I} - \vec{S}_v^{l} \vec{S}_h^{m})$, and $\vec{S}_v^{l}$ and $\vec{S}_h^{m}$ denote vertical and horizontal shifts of $\HR$ weighted by $\alpha_0 \in ]0; 1]$ over a $(2P + 1) \times (2P + 1)$ window. $\WeightsAMat = \text{diag}(\WeightsAFun _1, \ldots, \WeightsAFun_{N})$ with $\WeightsAFun_i \in [0; 1]$ are spatially adaptive weights to control the influence of the prior for edge preserving regularization.

\subsection{Non-Linear Least-Squares Estimation}
\label{sec:NonLinearLeastSquaresEstimation}

Since our energy function in \eqref{eqn:jointEnergyFunction} is non-linear \wrt $\MotionParams$, we propose a non-linear least-squares estimation of $\HR$ and $\MotionParams$. We first derive the least-squares approximation of \eqref{eqn:RegTerm} as:
\begin{equation}
	\left| \left| \vec{S} \HR \right| \right|_1 
	\approx \HR^\top \vec{L} \HR
	= \HR^\top \left( \sum_{l = -P}^P \sum_{m = -P}^P \vec{L}^{l,m} \right) \HR,
\end{equation}
where $\vec{L}^{l,m} = (\vec{S}^{l,m})^\top \vec{U}^{l,m} \vec{S}^{l,m}$. Here, $\vec{U}^{l,m}$ is a diagonal matrix associated with the shift $(l,m)$ and is assembled by $U^{l,m}_{ii} = (\max(|z_i|, \tau))^{-1}$, where $\vec{z} = \vec{S}^{l,m} \HR$ and $\tau$ is a small threshold ($\tau = 10^{-2})$ to avoid numerical instabilities close to zero. Next, we reformulate \eqref{eqn:jointEnergyFunction} as the least-squares term:
\begin{equation}
	\label{eqn:leastSquaresReformulation}
	E(\HR, \MotionParams) =
	\left| \left|
	\begin{pmatrix}
		\WeightsBMat^{\frac{1}{2}} \vec{r}(\HR, \MotionParams),
		& \sqrt{\lambda} \WeightsAMat^{\frac{1}{2}} \vec{L}^{\frac{1}{2}} \HR
	\end{pmatrix}^\top \right|\right|_2^2,
\end{equation} 
where $\vec{r}(\HR, \MotionParams) = \LR - \SystemMat(\MotionParams)\HR$ is the residual error. Accordingly, starting from an initial guess $\IterationIdx{\HR}{0}$ and $\IterationIdx{\MotionParams}{0}$, both parameter sets are iteratively updated according to $\IterationIdx{\HR}{t} = \IterationIdx{\HR}{t-1} + \Delta\HR$ and $\IterationIdx{\MotionParams}{t} = \IterationIdx{\MotionParams}{t-1} + \Delta \MotionParams$ for $t \geq 1$. The updates $\Delta\HR$ and $\Delta \MotionParams$ at iteration $t$ are derived by a linear approximation of \eqref{eqn:leastSquaresReformulation} and are determined as the minimum of the least-squares term:
\vspace{-0.5em}
\begin{align}
	\label{eqn:leastSquaresEquationSystem}
	&\left| \left|
	\IterationIdx{\vec{P}}{t} 
		\begin{pmatrix}
			\Delta \MotionParams \\
			\Delta \HR
		\end{pmatrix}
	 - \IterationIdx{\vec{f}}{t}
	 \right| \right|_2^2,\\
	\text{where}~\IterationIdx{\vec{P}}{t} &=
		\begin{pmatrix}
			(\IterationIdx{\WeightsBMat}{t})^\frac{1}{2} \vec{J}(\HR, \MotionParams) & (\IterationIdx{\WeightsBMat}{t})^\frac{1}{2} \SystemMat(\MotionParams)\\
			\vec{0} & \sqrt{\lambda} (\IterationIdx{\WeightsAMat}{t})^{\frac{1}{2}}\vec{L}^{\frac{1}{2}}
		\end{pmatrix}, \\
	\text{and}~\IterationIdx{\vec{f}}{t} &=
		\begin{pmatrix}
			(\IterationIdx{\WeightsBMat}{t})^\frac{1}{2} \vec{r(\HR, \MotionParams)} \\
			-\sqrt \lambda (\IterationIdx{\WeightsAMat}{t})^{\frac{1}{2}} \vec{L}^{\frac{1}{2}} \HR
		\end{pmatrix}.
\end{align}
$\vec{J}(\HR, \MotionParams) = \SamplingMat\BlurMat \frac{\partial(\MotionMat(\MotionParams)\HR)}{\partial \MotionParams}$ is the Jacobian of the system matrix $\SystemMat(\MotionParams)\HR$ \wrt $\MotionParams$. The derivative of $\MotionMat(\MotionParams)$ \wrt $\MotionParams$ is computed numerically using bilinear interpolation \cite{He2007}. 

For confidence-aware optimization, we estimate the observation weights $\IterationIdx{\WeightsBMat}{t}$ from $\vec{r} = \LR - \SystemMat(\IterationIdx{\MotionParams}{t-1}) \IterationIdx{\HR}{t-1}$ according to:
\begin{equation}
	\label{eqn:weightingFunctionsObservations}
	\IterationIdx{\WeightsBFun}{t}_i =
	\begin{cases}
		1  
			& \mbox{if } |r_i| \leq \IterationIdx{\NoiseStd}{t} \\
		\IterationIdx{\NoiseStd}{t} / |r_i|
			& \text{otherwise}
	\end{cases},
\end{equation}
and WBTV weights $\IterationIdx{\WeightsAMat}{t}$ from $\vec{z} = \vec{S}\IterationIdx{\HR}{t-1}$ according to:
\begin{equation}
	\label{eqn:weightingFunctions}
	\IterationIdx{\WeightsAFun}{t}_i = 
	\begin{cases}
		1  
			& \mbox{if } |z_i| \leq \IterationIdx{\PriorStd}{t} \\
		p \left( \IterationIdx{\PriorStd}{t} / |z_i| \right)^{1-p}
			& \text{otherwise}
	\end{cases},
\end{equation}
where $p \in [0,1]$ denotes a sparsity parameter. The scale parameters $\IterationIdx{\NoiseStd}{t} = 1.4826 \cdot \mathrm{mad}(\LR - \SystemMat(\IterationIdx{\MotionParams}{t-1}) \IterationIdx{\HR}{t-1}, \IterationIdx{\WeightsBMat}{t-1})$ and $\IterationIdx{\PriorStd}{t} = \mathrm{mad}(\vec{S}\IterationIdx{\HR}{t-1}, \IterationIdx{\WeightsAMat}{t-1})$ are determined via the weighted median absolute deviation (MAD) rule under the weights $\IterationIdx{\WeightsAMat}{t-1}$ and $\IterationIdx{\WeightsBMat}{t-1}$ as derived in \cite{Kohler2016}.

\begin{algorithm}[!t]
	\small
	\caption{Confidence-aware Levenberg-Marquardt optimization}
	\begin{algorithmic}[1]
		\STATE Initialize $\IterationIdx{\HR}{0}$, $\IterationIdx{\MotionParams}{0}$ and $t = 1$
		 \WHILE{Convergence criterion not fulfilled}
				\STATE Get $\IterationIdx{\WeightsBMat}{t}$ and $\IterationIdx{\WeightsAMat}{t}$ from ($\IterationIdx{\HR}{t-1},\IterationIdx{\MotionParams}{t-1})$ according to \eqref{eqn:weightingFunctionsObservations} and \eqref{eqn:weightingFunctions}
			 	\STATE Estimate $\IterationIdx{\mu}{t}$ from \eqref{eqn:muSearch} with search range $[\log \mu_l; \log \mu_u]$
			 	\STATE Estimate $\Delta \HR$ and $\Delta \MotionParams$ with $\mu = \IterationIdx{\mu}{t}$ 
			according to \eqref{eqn:lmUpdateRule}
				\STATE Update $\IterationIdx{\HR}{t} = \IterationIdx{\HR}{t-1} + \Delta\HR$ and $\IterationIdx{\MotionParams}{t} = \IterationIdx{\MotionParams}{t-1} + \Delta \MotionParams$ 
				\STATE Set $t \leftarrow t + 1$ and proceed with next iteration 
			\ENDWHILE 
	\end{algorithmic}
	\label{alg:lmAlgorithm}
\end{algorithm}

\subsection{Levenberg-Marquardt Iterations}
\label{sec:LevenbergMarquardtIterations}

Related SR algorithms formulated via least-squares estimation \cite{He2007, Zhang2012} employ Gauss-Newton iterations to iteratively minimize \eqref{eqn:leastSquaresReformulation}. However, the convergence of a Gauss-Newton scheme relies on an initial solution close to the desired global minimum, which is hard to guarantee under practical conditions, e.\,g. if the initial estimate is affected by outliers. To alleviate this issue, we propose Levenberg-Marquardt iterations, which combines the Gauss-Newton method with gradient descent in our method summarized in Algorithm \ref{alg:lmAlgorithm}. 
Instead of solving \eqref{eqn:leastSquaresEquationSystem}, we compute the updates $\Delta \MotionParams$ and $\Delta \HR$ in our Levenberg-Marquardt iteration scheme according to:
 \begin{equation}
	\label{eqn:lmUpdateRule}
	\begin{pmatrix} \Delta \MotionParams \\ \Delta \HR \end{pmatrix}
	= \Big[ (\IterationIdx{\vec{P}}{t})^\top \IterationIdx{\vec{P}}{t} 
	+ \mu \cdot \text{diag} \left( (\IterationIdx{\vec{P}}{t})^\top \IterationIdx{\vec{P}}{t} \right) \Big]^{-1} 
	(\IterationIdx{\vec{P}}{t})^\top \IterationIdx{\vec{f}}{t},
\end{equation}
where $\mu$ is the damping factor to control the contributions of gradient descent ($\mu \gg 0$) and Gauss-Newton ($\mu = 0$). The system in \eqref{eqn:lmUpdateRule} is solved by the conjugate gradient (CG) method with $T_{\text{cg}}$ iterations in our inner optimization loop to avoid a direct inversion of $\IterationIdx{\vec{P}}{t}$. In our outer optimization loop, we use $T_{\text{lm}}$ Levenberg-Marquardt iterations to minimize \eqref{eqn:leastSquaresReformulation}.

The performance of the Levenberg-Marquardt iterations in \eqref{eqn:lmUpdateRule} is highly depended on the damping parameter $\mu$. The common damping approach adaptively selects $\mu$ from only two candidates per iteration \cite{Marquardt1963}, which may lead to inadequate adaptations between gradient descend and Gauss-Newton. In contrast to this approach, we propose to select the damping parameter adaptively per iteration as $\IterationIdx{\mu}{t}$ to minimize the confidence weighted residual error $\vec{r}(\HR, \MotionParams)$. This leads to the one-dimensional optimization problem:
\begin{equation}
	\label{eqn:muSearch}
	\IterationIdx{\mu}{t} = \arg \min_{\mu} 
	\left| \left| (\IterationIdx{\WeightsBMat}{t})^{\frac{1}{2}} \left[\LR - \SystemMat \left ( \MotionParams(\mu) \right) \HR(\mu) \right] \right| \right|_2^2,
\end{equation}
where $\HR(\mu)$ and $\MotionParams(\mu)$ are the HR image and the motion parameters obtained from \eqref{eqn:lmUpdateRule} with damping factor $\mu$. To make this parameter selection tractable, we approximate \eqref{eqn:muSearch} by a one-dimensional search with $T_{\mu}$ iterations over the logarithmic scaled range $[\log \mu_l; \log \mu_u]$ in our algorithm. Subsequently, we solve \eqref{eqn:lmUpdateRule} with the optimal parameter $\mu = \IterationIdx{\mu}{t}$.   

\section{Experiments and Results}
\label{sec:ExperimentsAndResults}

This section studies the performance of the proposed method on real and simulated data. We compared our method to the confidence-aware two-stage algorithm based on iteratively re-weighted minimization (IRWSR) \cite{Kohler2016} and the joint motion estimation and SR using Gauss-Newton iterations (JMSR) \cite{He2007}. Throughout all experiments, we set the WBTV parameters to $P = 2$, $\alpha_0 = 0.5$ and $p = 0.5$. For iterative minimization, we chose $T_{\text{lm}} = 25$, $T_{\text{cg}} = 25$ and $T_{\mu} = 5$ with the damping parameter search range $\log\mu_l = -4$ and $\log\mu_u = 4$. The regularization weights were selected on one training sequence per dataset using a grid search to enable fair comparisons.
\begin{figure}[!t]
	\centering
	\subfloat{\includegraphics[width=0.46\textwidth]{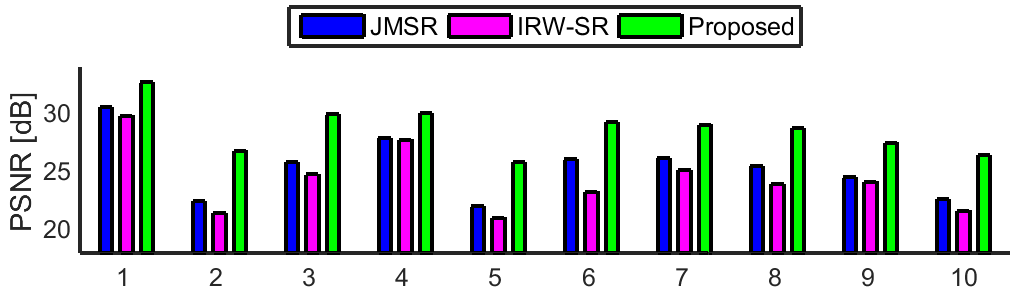}}\\[-0.25ex]
	\subfloat{\includegraphics[width=0.46\textwidth]{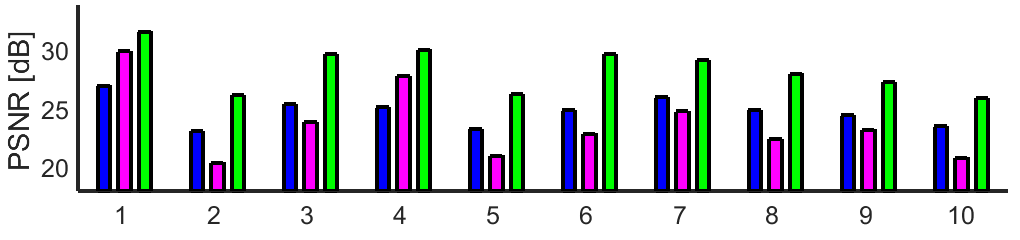}.png}
	\centering
	\caption{Mean PSNR over ten simulated datasets with ten randomly generated image sequences per dataset. We compared SR under inaccurate motion estimation without outliers (top row) and with outliers due to invalid pixels (bottom row).}
	\label{fig:bars}
\end{figure}
\begin{figure*}[!t]
	\centering
	\subfloat{\includegraphics[width=0.24\textwidth]{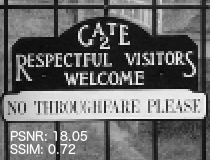}}~
	\subfloat{\includegraphics[width=0.24\textwidth]{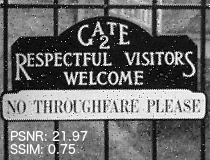}}~
	\subfloat{\includegraphics[width=0.24\textwidth]{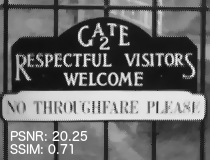}}~
	\subfloat{\includegraphics[width=0.24\textwidth]{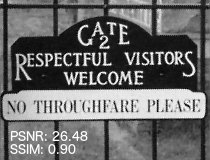}}\\[-1.4ex]
	\setcounter{subfigure}{0}
	\subfloat[Original]{\includegraphics[width=0.24\textwidth]{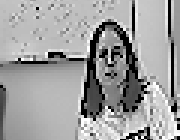}}~
	\subfloat[JMSR \cite{He2007}]{\includegraphics[width=0.24\textwidth]{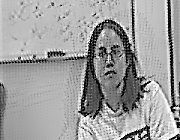}}~ 
	\subfloat[IRWSR \cite{Kohler2016}]{\includegraphics[width=0.24\textwidth]{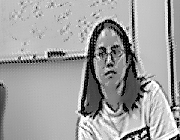}}~
	\subfloat[Proposed]{\includegraphics[width=0.24\textwidth]{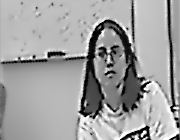}}
	\caption{SR reconstructions provided by the competing algorithms on simulated and real datasets. Top row: Simulated dataset with uncertainties in the initial motion parameters and outliers due to invalid pixels generated by salt-and-pepper noise. Bottom row: \textit{Emily} dataset with outliers due to non-rigid motion across the LR frames related to movements of the head.}
	\label{fig:parrots_alpaca}
\end{figure*}

\begin{figure}
	\scriptsize
    \centering
		\includegraphics[width=0.25\textwidth]{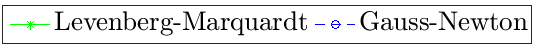}\\
    \newlength\figureheight
    \newlength\figurewidth
    \setlength\figureheight{2.1cm}
    \setlength\figurewidth{3.0cm}
%
%
\begin{tikzpicture}

\begin{axis}[%
width=\figurewidth,
height=\figureheight,
scale only axis,
separate axis lines,
every outer x axis line/.append style={white!15!black},
every x tick label/.append style={font=\color{white!15!black}},
xmin=1,
xmax=20,
every outer y axis line/.append style={white!15!black},
every y tick label/.append style={font=\color{white!15!black}},
ymin=24.5,
ymax=31.5,
ylabel={PSNR [dB]},
xlabel={Iteration},
legend style={at={(0.5,1.03)},anchor=south,legend columns=2,draw=white!15!black,fill=white,legend cell align=left}
]
\addplot [color=blue,dashed,mark=o,mark options={solid},mark size=1.3,forget plot]
  table[row sep=crcr]{1	25.1122294642163\\
2	26.7991172448951\\
3	27.752030907878\\
4	29.0665237759133\\
5	29.7226632132648\\
6	30.2640741285465\\
7	30.3008270001308\\
8	30.6780671386111\\
9	30.7364767483629\\
10	30.7598511352928\\
11	30.6598107159358\\
12	30.7015959146735\\
13	30.6253463210254\\
14	30.7873194952826\\
15	30.5623768132296\\
16	30.6722114208254\\
17	30.6395169705374\\
18	30.5776325696071\\
19	30.6136015623233\\
};
\addplot [color=green,solid,mark=x,mark options={solid},mark size=1.3,forget plot]
  table[row sep=crcr]{1	25.1121837604039\\
2	26.7986001582893\\
3	27.7515421544632\\
4	28.6648030216358\\
5	29.5587912457338\\
6	30.1960775505398\\
7	30.6413326872265\\
8	30.6606862563564\\
9	30.7298358988162\\
10	30.7428790416208\\
11	30.9790332119205\\
12	30.9306807724652\\
13	31.0126277326471\\
14	31.0866344156878\\
15	31.0866528841067\\
16	31.0866695597967\\
17	31.0866862319493\\
18	31.0882924860147\\
19	31.0883022872424\\
};
\end{axis}
\end{tikzpicture}%
\quad
%
%
\begin{tikzpicture}

\begin{axis}[%
width=\figurewidth,
height=\figureheight,
scale only axis,
separate axis lines,
every outer x axis line/.append style={white!15!black},
every x tick label/.append style={font=\color{white!15!black}},
xmin=1,
xmax=20,
xlabel={Iteration},
every outer y axis line/.append style={white!15!black},
every y tick label/.append style={font=\color{white!15!black}},
ymin=24.5,
ymax=31.5,
ylabel={PSNR},
legend style={at={(0.5,1.03)},anchor=south,legend columns=2,draw=white!15!black,fill=white,legend cell align=left}
]
\addplot [color=green,solid,mark=x,mark options={solid},mark size=1.3,forget plot]
  table[row sep=crcr]{
1	26.2464803084492\\
2	26.7916325676117\\
4	27.4285289403273\\
5	27.5222037163289\\
6	27.5345654614537\\
7	27.629002760083\\
8	27.6488003398717\\
9	27.7034770636676\\
10	27.7034809213808\\
11	27.7034843078079\\
12	27.7034878200462\\
13	27.7034913378664\\
14	27.7034948541908\\
15	27.7034983719383\\
16	27.7035018856259\\
17	27.7035053964643\\
18	27.7035089034077\\
19	27.70351240683\\
};
\addplot [color=blue,dashed,mark=o,mark options={solid},mark size=1.3,forget plot]
  table[row sep=crcr]{1	24.8837133738916\\
2	25.4990795440115\\
3	25.7477450281175\\
4	26.1517235578444\\
5	26.1476668350712\\
6	26.5204159264985\\
7	26.3325990503675\\
8	26.5308333886108\\
9	26.2657411761878\\
10	26.4208656044542\\
11	26.1838570582405\\
12	26.3820705197855\\
13	26.0824156660364\\
14	26.1619716376507\\
15	25.9729663826569\\
16	26.0614488347627\\
17	25.9448848106759\\
18	26.1893945052941\\
19	26.194196453598\\
};
\end{axis}
\end{tikzpicture}%
    \caption{PSNR over the iterations of our algorithm using Gauss-Newton and the proposed Levenberg-Marquardt iterations. Left: Iterations without outliers. Right: Iterations in the presence of outliers due to invalid pixels.}
    \label{fig:lm_vs_gn}
\end{figure}
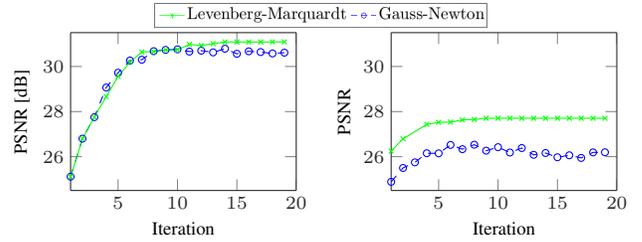 

For our quantitative evaluation, we generated synthetic datasets from ten HR images taken from the LIVE database \cite{Sheikh2016}. The LR data was simulated from the HR images according to \eqref{eqn:imageFormationModel} using a decimation factor of 2, a Gaussian PSF of width $6 \cdot \PSFWidth$ ($\PSFWidth = 0.5$), and uniformly distributed translations ($-2$ to $2$\,px) and rotations ($-1^\circ$ to $1^\circ$). Each frame was corrupted by additive, Gaussian  noise with standard deviation $\NoiseStd = 0.025$. The fidelity of the SR reconstruction to the ground truth data was assessed by the peak-signal-to-noise ratio (PSNR) in decibels (dB) as well as structural similarity (SSIM). We simulated sequences consisting of $K = 12$ frames with two scenarios: 1) The initial motion parameters were corrupted by uniformly distributed errors for the translation ($-0.3$ to $+0.3$\,px) and the rotation ($-0.005^\circ$ to $+0.005^\circ$) to simulate realistic accuracies of motion estimation in a baseline experiment. 2) We corrupted two frames per sequence by salt-and-pepper noise with noise level $\nu = 0.075$ that denotes the fraction of invalid pixels to simulate outliers in the input frames. Fig. \ref{fig:bars} depicts the PSNR for ten datasets with ten randomly generated realizations of these experiments per dataset. We observed that on the one hand JMSR compensated uncertainties in the motion parameters but deteriorates in the presence of outliers. On the other hand, IRWSR was robust regarding outliers but could not correct uncertainties in the initial motion parameters. In contrast to these methods, our confidence-aware algorithm compensated both effects. On average, our method outperformed JMSR by 3.2 dB in the absence of outliers and by 3.0 dB in presence of outliers. See Fig. \ref{fig:parrots_alpaca} (top row) for a qualitative comparison along with the PSNR and SSIM measures of the HR images. Here, our method corrected motion estimation uncertainties and was insensitive to invalid pixels. 

In order to prove the benefit of the proposed Levenberg-Marquardt optimization, we studied the convergence of our algorithm using the proposed iteration scheme in comparison to Gauss-Newton iterations similar to \cite{He2007}. Fig. \ref{fig:lm_vs_gn} depicts this comparison on a simulated dataset generated with motion estimation uncertainties and outliers due to salt-and-pepper noise. In both scenarios, the proposed Levenberg-Marquardt scheme converged within the first ten iterations and outperformed simple Gauss-Newton iterations.

Finally, we evaluated our method on real image sequences from the MDSP dataset \cite{Farsiu2016}. Fig. \ref{fig:parrots_alpaca} (bottom row) shows a comparison of the competing SR approaches on the \textit{Emily} sequence, where the initial motion estimation was performed on the LR frames using the method in \cite{Evangelidis2008}. This sequence follows a translational motion in the first part and non-rigid motion related to head movements in the second part. This deviation of the actual motion to the rigid model caused outliers that cannot be compensated by JMSR and resulted in artifacts in the estimated HR image. The IRWSR method was able to compensate these outlier frames but could not refine the initial motion estimation for the remaining frames. Our method combines outlier removal with a refinement of the initial motion parameters resulting in less artifacts. 

\section{Conclusion}
\label{sec:Conclusion}

In this paper, we introduced a new joint motion estimation and SR algorithm. Unlike related methods, our algorithm is based on a confidence-aware formulation to consider outliers and space variant noise in the image formation. We developed Levenberg-Marquardt optimization that jointly estimates motion parameters, the unknown HR image and model confidence weights. In our experiments on real and simulated data, our algorithm outperformed state-of-the-art two-stage SR reconstruction as well as joint motion estimation and SR by combining the advantages of both approaches. In particular, the proposed algorithm increases robustness regarding inaccurate motion estimation and outlier observations.

In our future work, we aim at extending our model to more general types of motion, e.\,g. affine transformations.

\bibliographystyle{IEEEbib}
\bibliography{refs}

\end{document}